\documentclass{article}

 \usepackage{neurips_2025} 

\usepackage[utf8]{inputenc} 
\usepackage[T1]{fontenc}    
\usepackage{hyperref}       
\usepackage{url}  
\usepackage{booktabs}       
\usepackage{amsfonts}       
\usepackage{nicefrac}       
\usepackage{microtype}      
\usepackage{xcolor}         
\usepackage{natbib}         
\usepackage{amsmath}
\usepackage{amssymb}
\usepackage{graphicx} 
\usepackage{siunitx} 
\usepackage{listings} 
\usepackage{ragged2e} 
\usepackage{enumitem} 
\usepackage{algorithm} 
\usepackage{algorithmic} 

\title{Mastering Da Vinci Code: A Comparative Study of Transformer, LLM, and PPO-based Agents}

\author{%
  LeCheng Zhang \quad Yuanshi Wang \quad Haotian Shen \quad Xujie Wang \\
  Westlake College, Westlake University \\
  \texttt{\{zhanglecheng, wangyuanshi, shenhaotian,wangxujie\}@westlake.edu.cn}
}

\begin{document}

\maketitle

\begin{abstract}
The Da Vinci Code, a game of logical deduction and imperfect information, presents unique challenges for artificial intelligence, demanding nuanced reasoning beyond simple pattern recognition. This paper investigates the efficacy of various AI paradigms in mastering this game. We develop and evaluate three distinct agent architectures: a Transformer-based baseline model with limited historical context, several Large Language Model (LLM) agents (including Gemini, DeepSeek, and GPT variants) guided by structured prompts, and an agent based on Proximal Policy Optimization (PPO) employing a Transformer encoder for comprehensive game history processing. Performance is benchmarked against the baseline, with the PPO-based agent demonstrating superior win rates ($58.5\% \pm 1.0\%$), significantly outperforming the LLM counterparts. Our analysis highlights the strengths of deep reinforcement learning in policy refinement for complex deductive tasks, particularly in learning implicit strategies from self-play. We also examine the capabilities and inherent limitations of current LLMs in maintaining strict logical consistency and strategic depth over extended gameplay, despite sophisticated prompting. This study contributes to the broader understanding of AI in recreational games involving hidden information and multi-step logical reasoning, offering insights into effective agent design and the comparative advantages of different AI approaches. \thanks{Code available at: \url{https://github.com/zhang-lecheng/Davinci-Code-Agent/}}
\end{abstract}

\section{Introduction}
The central problem addressed in this research is the development and rigorous comparative analysis of artificial intelligence agents capable of achieving high-level performance in the Da Vinci Code. This game, while having relatively simple rules, epitomizes a class of problems requiring intricate logical deduction under conditions of imperfect information. Players must infer opponents' hidden tiles by synthesizing information from public cues (their own tiles, revealed opponent tiles, past guesses) and the game's strict tile ordering rules. Successfully navigating this environment demands not only logical prowess but also probabilistic reasoning and an understanding of opponent tendencies, making it a formidable challenge for AI systems.

Solving or achieving strong performance in Da Vinci Code is important because it pushes AI capabilities in several critical dimensions. Firstly, it serves as a valuable benchmark for AI in strategic reasoning within partially observable environments, moving beyond games of perfect information like Go \citep{silver2016mastering} or those where statistical exploitation is paramount, like Poker \citep{brown2019superhuman}. Success here implies an ability to handle uncertainty and perform multi-step, iterative deduction. Secondly, a comparative study of different AI paradigms—specifically, rule-based Transformers, general-purpose LLMs, and specialized Deep Reinforcement Learning (DRL) agents—in this domain can yield crucial insights. It allows us to understand their respective strengths, weaknesses, and computational trade-offs when applied to tasks demanding robust logical inference, precise state tracking, and adaptive long-term planning. Such insights are transferable to real-world problems involving complex decision-making with incomplete data, such as intelligence analysis or diagnostic reasoning.

The Da Vinci Code presents significant difficulties for AI due to a confluence of factors. The game state is partially observable, forcing agents to reason about distributions of hidden information rather than a single, known state. The core gameplay revolves around sequential, multi-step logical deduction that must adapt as new information is revealed; this requires maintaining and updating a complex belief state. The combinatorial nature of possible tile arrangements and deduction pathways, while not as astronomically large as Go, is substantial.
Naive methods often prove insufficient. For instance, a simple Transformer baseline, even if proficient at pattern matching on the current board state, fails if it lacks access to or the ability to meaningfully process the full game history. Such a model would struggle to infer constraints from past incorrect guesses or to model the narrowing set of possibilities for hidden tiles over time. Similarly, while powerful general-purpose LLMs \citep{openai2023gpt4, gemini2023, deepseek2024} demonstrate impressive zero-shot or few-shot reasoning, they may falter in Da Vinci Code due to several reasons: (a) maintaining strict adherence to the game's rigid sorting and deduction rules over many turns can be challenging; (b) performing consistently flawless multi-step logical inference, especially when probabilities are involved, is not their primary design strength without specialized fine-tuning or highly structured, iterative prompting frameworks (e.g., tree-of-thoughts); (c) their reasoning is not inherently grounded in the game's formal system, potentially leading to plausible but incorrect deductions, or an inability to explore the full deductive space optimally.

Prior AI research in complex card and tile games like Guandan \citep{pan2024mastering}, DouDizhu \citep{zha2021douzero}, and Mahjong \citep{li2020suphx}, while achieving superhuman feats, primarily relies on DRL and Monte Carlo methods. These methods, though powerful, often face challenges in sample efficiency, effective exploration in vast state-action spaces, and stable convergence, particularly for games with nuanced multi-agent interactions (cooperation/competition) or extremely large branching factors. For Da Vinci Code specifically, there is a dearth of prior work focusing on specialized AI agents. Applying general AI techniques without tailoring them to the game's deductive core poses limitations. Standard DRL approaches might struggle if the state representation does not adequately capture the logical constraints or if the reward shaping is insufficient to guide complex deductive strategies. As for LLMs, their primary limitation in this context, beyond potential rule adherence issues, is that their pre-training does not inherently equip them with the specialized deductive algorithms optimal for Da Vinci Code; they attempt to derive them on-the-fly, which can be inefficient or error-prone compared to a system explicitly learning or built with such logic.

This paper addresses these challenges by systematically developing, implementing, and evaluating three distinct classes of AI agents for Da Vinci Code:
\begin{enumerate}
    \item \textbf{A Transformer-based Baseline Model:} This agent, inspired by the success of Transformers \citep{vaswani2017attention} in sequence processing, is designed to predict opponent tiles based on the current public information. However, it operates without access to the full game history, serving as a benchmark for more sophisticated, history-aware approaches.
    \item \textbf{LLM-based Agents:} We leverage several prominent LLMs (Gemini, DeepSeek, GPT-4o, Qwen) as game-playing agents. These agents are guided by carefully engineered prompts that provide game rules, current state information (including parsed history of incorrect guesses), and strategic heuristics, aiming to elicit logical reasoning within the game's constraints.
    \item \textbf{A PPO-based Agent:} This agent, grounded in Proximal Policy Optimization (PPO) \citep{schulman2017proximal}, employs a Transformer encoder to process a comprehensive representation of the game history and current state. It learns a policy through extensive self-play, aiming to discover and refine effective deductive and guessing strategies.
\end{enumerate}
Our methodology involves not only the implementation of these agents but also a rigorous comparative performance analysis against the baseline and, implicitly, against human expert benchmarks. This allows us to elucidate the most effective AI strategies for Da Vinci Code and, by extension, for other deductive games with imperfect information. Our findings underscore the significant advantages of specialized policy learning achieved through DRL, particularly when compared to the more generalized reasoning capabilities of LLMs in this specific, logic-intensive domain.

\section{Related Work}
The development of AI capable of excelling in games with imperfect information and complex state spaces has been a long-standing goal. Early successes in games like Backgammon \citep{tesauro1995td}, which combined temporal difference learning with neural networks, paved the way for more recent breakthroughs in significantly more complex environments.

\paragraph{AI in Imperfect Information Games}
Poker, particularly No-Limit Texas Hold'em, has been a significant benchmark for AI in imperfect information settings. Systems like Libratus \citep{brown2018superhuman} and Pluribus \citep{brown2019superhuman} demonstrated superhuman performance in two-player and multi-player variants, respectively. These systems often rely on techniques such as Counterfactual Regret Minimization (CFR) and its variants, coupled with sophisticated abstraction and solving techniques for large game trees. While highly successful, these methods are often tailored to the specific structure of Poker, which involves elements of bluffing, betting strategy, and hand strength evaluation that are distinct from the primarily deductive nature of Da Vinci Code.

\paragraph{AI in Asian Card and Tile Games}
Recent research into popular Asian games, which often feature complex rules, hidden information, and multi-agent dynamics, has highlighted the efficacy of DRL and Monte Carlo Tree Search (MCTS).
\citet{zha2021douzero} introduced DouZero for DouDizhu, a three-player card game involving shifting alliances (one "landlord" vs. two "peasants") and a vast action space. DouZero, which leveraged DRL with improved Monte Carlo methods (specifically, avoiding explicit MCTS and using deep neural networks with action encoding and parallel actors), achieved superhuman performance, outperforming existing AI programs and top human players. The authors emphasized that their approach demonstrated how classic Monte Carlo methods, when properly augmented with deep neural networks and tailored to the action space, can achieve strong results even in domains previously considered too complex for them.
For Mahjong, a game with rich hidden information and intricate scoring rules, \citet{li2020suphx} developed Suphx. This AI mastered Mahjong using DRL augmented with novel techniques such as global reward prediction (to handle sparse and delayed rewards), oracle guiding (to aid exploration during pre-training), and runtime policy adaptation (to adjust to opponents' styles). Suphx outperformed most top human players on the Tenhou platform.
More recently, \citet{pan2024mastering} proposed GuanZero for Guandan, a four-player, two-versus-two card game with an exceptionally large action space (estimated up to $10^5$ legal moves). Their framework also utilized MCTS guided by deep neural networks, with a significant contribution being a carefully designed neural network encoding scheme to effectively regulate agent behavior and manage the vast search space.
These studies collectively show that DRL and MCTS are powerful tools, but significant challenges remain. These include the efficiency of policy optimization, effective exploration in massive state-action spaces, robust behavior regulation, and handling nuanced cooperative/competitive dynamics. Da Vinci Code, while also an imperfect information game, differs in its core mechanics: it emphasizes pure logical deduction from a sequence of revealed information and tile ordering constraints, rather than the complex card-play strategies, partnerships, or intricate scoring systems central to Guandan, DouDizhu, or Mahjong. This focus on deduction may require different architectural or algorithmic emphases.

\paragraph{Large Language Models in Games}
LLMs, such as those from OpenAI (e.g., GPT-4 \citep{openai2023gpt4}), Google (e.g., Gemini \citep{gemini2023}), DeepSeek \citep{deepseek2024}, and Alibaba (e.g., Qwen \citep{bai2023qwen}), built upon the Transformer architecture \citep{vaswani2017attention}, have demonstrated remarkable emergent capabilities in natural language understanding, generation, and few-shot reasoning across a wide array of tasks. Their application to complex games is an active and evolving area of research. While LLMs can understand game rules from descriptions and generate plausible moves or strategies via prompting, questions remain regarding their ability to: (1) maintain long-term strategic coherence over extended gameplay; (2) strictly adhere to all explicit and implicit game rules without "hallucinating" illegal moves or states; (3) perform robust, multi-step logical inference under uncertainty, especially when it requires precise symbolic manipulation or probabilistic calculations not directly embedded in their training data; and (4) efficiently explore the game tree or learn optimal policies without explicit game-playing experience. This paper explores their direct application as game-playing agents for Da Vinci Code through carefully designed prompting strategies.

Our work contributes to this landscape by focusing on a deduction-centric game, Da Vinci Code, and systematically comparing a history-aware RL agent (PPO) against various state-of-the-art LLMs and a simpler, history-limited Transformer baseline. This comparison aims to shed light on the most suitable AI paradigms for games where logical inference is paramount.

\section{Methodology}
We developed and evaluated three distinct categories of AI agents for the Da Vinci Code game: a Transformer-based baseline model, several LLM-based agents, and a PPO-trained agent. Each agent interacts with a custom-built game environment.

\subsection{Game Environment (\texttt{DaVinciCodeGameEnvironment})}
The game logic is encapsulated within the \texttt{DaVinciCodeGameEnvironment} class, implemented in \texttt{env.py}. This environment adheres to the standard Da Vinci Code rules for two players.
\begin{itemize}
    \item \textbf{Initialization and Deck:} The game is initialized for two players. The deck comprises 24 unique numbered tiles (B0-B11, W0-W11) and two joker tiles (B-, W-), often represented as hyphens. Each player is dealt an initial hand (defaulting to 4 tiles). A crucial aspect of our joker implementation is that their effective numerical values (\texttt{black\_joker}, \texttt{white\_joker}) are randomized floats at the start of each game. This ensures their relative order can vary while still adhering to the rule that jokers can represent any value, and their specific value is determined upon placement for sorting relative to other tiles.
    \item \textbf{State Representation (\texttt{\_get\_state}):} The environment provides a comprehensive state dictionary for the current player. This includes: the \texttt{current\_player} ID; their \texttt{hand} of tiles; a boolean list \texttt{revealed\_self} indicating which of their own tiles are face-up; a list \texttt{opponent\_hand\_visible} representing the opponent's hand from the current player's perspective (e.g., showing revealed cards like 'B7', or only the color of hidden cards like 'W?'); a boolean list \texttt{opponent\_revealed} detailing the revealed status of each of the opponent's cards; the current \texttt{deck\_size}; the \texttt{drawn\_card} by the current player (if any); a flag \texttt{can\_guess\_again} if a previous guess was correct; a list of \texttt{legal\_actions}; and game termination flags \texttt{game\_over} and \texttt{winner}.
    \item \textbf{Actions (\texttt{\_get\_legal\_actions}, \texttt{step}):} The environment supports two primary actions: \texttt{('place',)} to place the drawn card into the hand (its revealed status depends on whether the player \texttt{can\_guess\_again}), and \texttt{('guess', position, card\_value)} to guess the card at a specific \texttt{position} in the opponent's hand to be a certain \texttt{card\_value}.
    \item \textbf{Card Handling and Ordering (\texttt{card\_value}, \texttt{insert\_card}):} Tiles are always maintained in ascending order. Numerical value is the primary sort key. For tiles with the same numerical value, Black tiles are considered smaller than (i.e., placed to the left of) White tiles. Jokers are integrated into this order based on the specific value they represent at the time of placement, which is inferred by their randomized float value relative to adjacent tiles. The \texttt{insert\_card} method ensures this order is maintained when new cards are added to a hand.
    \item \textbf{Reward Structure (\texttt{step}):} The reward function is designed to guide learning:
        \begin{itemize}
            \item Winning the game yields a significant positive reward: +3.0.
            \item Losing the game (opponent wins) incurs a significant negative reward: -3.0.
            \item A correct guess provides a small positive reward: +0.2.
            \item An incorrect guess, which forces the player to reveal one of their own tiles, results in a small negative reward: -0.5.
            \item Placing a card that remains hidden (after a correct guess and choosing to end the guessing phase) is associated with the +0.2 reward from the preceding correct guess or can be seen as neutral if the agent simply passes the turn.
            \item Placing a card that is revealed (either at the start of a turn by choice, or as a penalty for an incorrect guess) is implicitly tied to the penalty of an incorrect guess (-0.5) or may offer no direct reward if chosen voluntarily without a prior successful guess.
            \item Invalid actions or internal environment errors can lead to a -3.0 reward and immediate game termination to penalize undesirable agent behavior.
        \end{itemize}
    \item \textbf{Game History Tracking (\texttt{history}):} A chronological list of strings (\texttt{history}) logs all major game events, such as draws, guesses (with outcomes), and placements. This history is crucial for agents that learn from or reason over past events.
\end{itemize}

\subsection{Baseline Model}
The baseline model utilizes a Transformer architecture \citep{vaswani2017attention}, a cornerstone of many modern NLP and sequence processing tasks, to predict the opponent's hand token by token.
\paragraph{Input Representation:} It processes a structured input string that describes the current game state from the active player's perspective. This string includes: (a) the player's own hand, with all tiles explicitly listed and sorted; (b) the opponent's hand, where revealed tiles show their full value (e.g., `[B7]`) and unrevealed tiles show only their color (e.g., `[W?]`) or color along with a list of values that the tile is known *not* to be, based on previous incorrect guesses for that specific slot (e.g., `[B?:!3 5]`).
\paragraph{Prediction Task:} The model is trained to sequentially predict the actual hidden values of the opponent's tiles.
\paragraph{Limitations:} A primary limitation of this baseline is its restricted access to the full game history. It primarily reasons based on the current snapshot of public information and explicit negative constraints from prior guesses on specific slots. It does not, for example, track the overall sequence of play, the number of cards drawn, or infer probabilities based on the dwindling deck, thus missing out on deeper strategic inferences available to more history-aware agents.

\subsection{LLM-based Agents}
We employed several state-of-the-art Large Language Models as game-playing agents, including Gemini-2.5-Flash-Preview-05-20 \citep{gemini2023}, various DeepSeek models (V3 and R1) \citep{deepseek2024}, GPT-4o \citep{openai2024gpt4o}, an unspecified "O3" model, and Qwen3-235B-A22B \citep{bai2023qwen}. The interaction with these models was facilitated through carefully engineered prompts, as exemplified by the Gemini agent implementation in \texttt{gemini.py}.
\paragraph{Input String Construction for LLMs:} The prompt provided to the LLMs is more comprehensive than the baseline's input. It includes:
\begin{enumerate}
    \item The current player's complete hand, fully visible to the LLM.
    \item The opponent's hand, formatted to show revealed tiles explicitly and hidden tiles as `[Color?:!guesses]`. The `!guesses` component is dynamically populated by parsing the game history to identify values previously guessed incorrectly *by the LLM agent itself* for that specific tile position. This allows the LLM to avoid repeating its own past mistakes for a given slot.
    \item The specific tile drawn by the current player at the start of their turn.
    \item A condensed summary of the last few (e.g., 6) game actions to provide immediate historical context.
\end{enumerate}
\paragraph{Prompting Strategy (Gemini Example):} A detailed system prompt was designed to guide the Gemini model. This prompt:
\begin{itemize}
    \item Assigns the role of an "expert Da Vinci Code player."
    \item Mandates output in a strict JSON format (e.g., \texttt{\{"action":"guess","position":2,"card":"B5"\}} or \texttt{\{"action":"place"\}}).
    \item Emphasizes strategic priorities, such as prioritizing a 'guess' action if unrevealed opponent tiles exist.
    \item Provides a summary of the input format it will receive.
    \item Outlines key game rules and facts (deck composition, tile ordering logic).
    \item Suggests a specific decision heuristic:
        \begin{enumerate}
            \item For each unrevealed opponent slot, compute the set of possible values consistent with known information (revealed neighbors, ordering rules, previous incorrect guesses for that slot, and globally revealed/held tiles).
            \item Select the slot with the smallest set of remaining candidate values (breaking ties by choosing the leftmost such slot).
            \item Guess the median value from this candidate set.
        \end{enumerate}
\end{itemize}
The user message then appends the dynamically generated game state string, the drawn card, and the list of currently legal actions as per the game environment.
\paragraph{Output Parsing and Fallback Mechanism:} The LLM's JSON response is parsed to extract the intended action. If the response is not valid JSON, or if the proposed action is not among the legal actions provided by the environment (indicating a rule violation or misunderstanding by the LLM), a fallback mechanism is triggered. This typically involves defaulting to a 'place' action if permissible, or, as a last resort, making a random legal guess to ensure game progression.

\subsection{PPO-based Agent}
The Proximal Policy Optimization (PPO) based agent \citep{schulman2017proximal} is designed to learn an optimal policy through self-play, leveraging comprehensive game history. Its architecture and training regimen are detailed in \texttt{test\_ppo.py}.
\paragraph{State Representation and Encoding:}
The transformation of raw game state into a suitable input for the neural network is a critical step, handled by the \texttt{convert\_env\_state\_to\_agent\_input} function.
\begin{itemize}
    \item \textbf{String Construction Rationale:} The game state is serialized into a string that includes: (a) the current player's hand (e.g., \texttt{Current-hand: B1 W5 B-}), (b) the opponent's publicly visible hand (e.g., \texttt{Opponent-hand: B? [W3] B?}), (c) the card drawn by the current player this turn, and (d) a processed representation of the game history. The history component is crucial; it's not just a raw log but is parsed to reflect a sequence of game events (Guess, Draw) with associated player IDs, positions, and card values/colors, normalized for the current player's perspective. This structured string aims to provide a rich, sequential context for the Transformer encoder.
    \item \textbf{Tokenization:} This string is then tokenized using a Byte Pair Encoding (BPE) tokenizer (\texttt{bpe\_tokenizer}) trained on a domain-specific vocabulary (\texttt{VOCAB}). This vocabulary includes all possible card names (e.g., 'B0', 'W-'), player identifiers ('Player0:', 'Player1:'), action types ('Guess', 'Draw'), positional markers ('pos0'-'pos13'), and structural elements ('Current-hand:', '\textbackslash n'). The vocabulary size is 64, with \texttt{PAD\_ID} explicitly defined (53).
    \item \textbf{Padding/Truncation:} The resulting sequence of token IDs is either padded with \texttt{PAD\_ID} or truncated to a fixed maximum history length (\texttt{MAX\_HISTORY\_LEN} = 256) to ensure consistent input dimensions for the neural network.
    \item \textbf{StateEncoder Network Architecture:} The tokenized \texttt{history\_ids\_tensor} and its corresponding \texttt{history\_padding\_mask\_tensor} are fed into the \texttt{StateEncoder} module. This module consists of:
        \begin{enumerate}
            \item An \texttt{nn.Embedding} layer to convert token IDs into dense vectors (embedding dimension 128).
            \item A separate \texttt{nn.Embedding} layer for positional encodings, capturing the sequential nature of the history.
            \item The token and positional embeddings are summed element-wise.
            \item These combined embeddings are then processed by a \texttt{nn.TransformerEncoder}. As per the \texttt{test\_ppo.py} implementation, this encoder defaults to 3 layers and 4 attention heads, with an input/model dimension of 128 and a feed-forward dimension typically twice that. Dropout (0.1) is applied within the Transformer layers for regularization.
            \item The final state representation is derived from the Transformer encoder's output sequence. Specifically, the output embedding corresponding to the \textit{first token} of the input sequence (analogous to a \texttt{[CLS]} token in BERT-like models) is taken as the aggregated representation of the entire game state and history (\texttt{history\_context\_encoded[:, 0, :]}). This fixed-size vector then serves as the input to the policy and value networks.
        \end{enumerate}
\end{itemize}
\paragraph{Policy and Value Networks:}
The state representation from the \texttt{StateEncoder} is passed to distinct Actor (policy) and Critic (value) networks, which share initial layers but have separate output heads.
\begin{itemize}
    \item \textbf{Actor Network (\texttt{Actor}):}
        The state vector is first processed by a shared Multi-Layer Perceptron (MLP), typically consisting of one linear layer projecting to 128 units followed by a ReLU activation function. The output of this shared MLP is then fed into the actor head: a single \texttt{nn.Linear} layer that outputs logits for each action in a flattened, discrete action space. This action space has $13 \times 26 + 1 = 339$ possible actions, corresponding to guessing each of 13 opponent tile positions with one of 26 card values, plus one action for 'placing' the drawn card. During action selection, a mask derived from the environment's legal actions is applied to these logits (setting illegal action logits to a very small number), and an action is sampled from the resulting \texttt{Categorical} distribution.
    \item \textbf{Critic Network (\texttt{Critic}):}
        This network also receives the state vector from its own instance of the \texttt{StateEncoder} (which is typically kept synchronized or shares weights with the actor's encoder during training). It uses a similar shared MLP structure. The critic head is a \texttt{nn.Linear} layer that outputs a single scalar value, $V(s)$, representing the estimated value (expected cumulative future reward) of the current state.
\end{itemize}
\paragraph{Action Space and Training Details:}
\begin{itemize}
    \item \textbf{Action Space Definition:} The agent interacts with the environment using a discrete, flattened action space of 339 actions. Index 338 is reserved for the \texttt{('place',)} action. Indices 0 through 337 map to guess actions, systematically encoded as \texttt{position\_index * num\_card\_values + card\_index}, where there are 13 potential guess positions and 26 unique card values (B0-B11, B-, W0-W11, W-).
    \item \textbf{Action Conversion Utilities:}
        \begin{itemize}
            \item \texttt{convert\_legal\_actions\_to\_masks}: This function translates the list of legal action tuples provided by the game environment into a boolean mask compatible with the 339-dimensional flat action space, enabling the actor to only consider valid moves.
            \item \texttt{convert\_actor\_action\_to\_env\_action}: Conversely, this function converts the flat action index sampled by the Actor network back into an environment-understandable action tuple (e.g., \texttt{('guess', pos, 'B5')}).
        \end{itemize}
    \item \textbf{PPO Training Regimen:}
        \begin{itemize}
            \item \textbf{Experience Collection:} Agent-environment interactions (state, action, reward, done, old log probability of action, value estimate $V(s_t)$) are stored in a \texttt{TrajectoryBuffer}. Data collection proceeds until \texttt{MAX\_TRAJECTORY\_LENGTH\_PER\_UPDATE} (16384 steps) are gathered.
            \item \textbf{Advantage Calculation:} Generalized Advantage Estimation (GAE) \citep{schulman2015high} is employed to compute advantages ($\hat{A}_t$) and returns ($G_t = \hat{A}_t + V(s_t)$) for updating the critic. Key GAE parameters are $\gamma$ (\texttt{GAMMA} = 0.999) for discounting future rewards and $\lambda$ (\texttt{GAE\_LAMBDA} = 0.95) for controlling the bias-variance trade-off in advantage estimation. Advantages are typically normalized before use.
            \item \textbf{Network Updates:} The policy (Actor) and value (Critic) networks are updated for \texttt{EPOCHS\_PER\_UPDATE} (3) iterations over the collected batch of trajectories, using mini-batches of size \texttt{BATCH\_SIZE\_PPO\_UPDATE} (2048).
            \item \textbf{Actor Loss:} The core of PPO is its clipped surrogate objective function: $L^{CLIP}(\theta) = \mathbb{E}_t [\min(r_t(\theta)\hat{A}_t, \text{clip}(r_t(\theta), 1-\epsilon, 1+\epsilon)\hat{A}_t)]$, where $r_t(\theta) = \frac{\pi_\theta(a_t|s_t)}{\pi_{\theta_{old}}(a_t|s_t)}$ is the probability ratio. An entropy bonus term, scaled by \texttt{ENTROPY\_COEFF} (0.01), is added to the loss to encourage exploration and prevent premature policy convergence. The clipping parameter $\epsilon$ is set to \texttt{PPO\_CLIP\_EPSILON} (0.2).
            \item \textbf{Critic Loss:} The value function is updated by minimizing the Mean Squared Error (MSE) between the Critic's predicted state values $V(s_t)$ and the calculated target returns $G_t$.
            \item \textbf{Optimizers:} The Adam optimizer \citep{kingma2014adam} is used for both actor and critic updates. The \texttt{actor\_optimizer\_instance} updates the parameters of its associated \texttt{StateEncoder} and the \texttt{Actor} network. Similarly, the \texttt{critic\_optimizer\_instance} updates its \texttt{StateEncoder} copy (if distinct) and the \texttt{Critic} network. The learning rate for the actor (and its encoder) is specified as 0.0005 in \texttt{test\_ppo.py}, while the critic's learning rate is 0.0003.
        \end{itemize}
\end{itemize}

\section{Experiments and Results}
We evaluated the performance of the developed agents against the Transformer-based baseline model. Table \ref{tab:model_performance} summarizes the win rates and 95\% confidence intervals.

\begin{table}[htbp]
\centering
\caption{Model Performance against Baseline Model (95\% CI)}
\label{tab:model_performance}
\begin{tabular}{l S[table-format=5.0] S[table-format=4.0] S[table-format=1.3] c}
\toprule
Model                     & {Games (n)} & {Wins} & {Win Rate (p)} & {Win Rate $\pm$ 95\% CI (\%)} \\
\midrule
\multicolumn{5}{l}{\textit{LLMs - Ous Category Models}} \\
Deepseek V3               & 100         & 20     & 0.200          & $20.0 \pm 7.8$             \\
GPT-4o                    & 100         & 17     & 0.170          & $17.0 \pm 7.4$             \\
\midrule
\multicolumn{5}{l}{\textit{LLMs - Other Models}} \\
O3                        & 95          & 35     & 0.368          & $36.8 \pm 9.7$             \\
Gemini-2.5-Flash-Preview-05-20      & 61          & 19     & 0.311          & $31.1 \pm 11.6$            \\
Deepseek R1               & 36          & 14     & 0.389          & $38.9 \pm 15.9$            \\
Qwen3-235B-A22B           & 29          & 8      & 0.276          & $27.6 \pm 16.3$            \\
\midrule
\multicolumn{5}{l}{\textit{PPO-based Model}} \\
PPO LM                    & 10000       & 5850   & 0.585          & $58.5 \pm 1.0$             \\
\midrule
\multicolumn{5}{l}{\textit{Human Experts}} \\
Human Expert (Median)     & 20          & 10     & 0.500          & $50.0 \pm 21.9$            \\
Human Expert (High)       & 25          & 16     & 0.640          & $64.0 \pm 18.8$            \\
\bottomrule
\end{tabular}
\small
\textit{Note: Some LLMs (O3, Gemini-2.5-Flash-Preview-05-20) were noted to occasionally produce outputs inconsistent with game logic.}
\end{table}

\begin{figure}[htbp]
    \centering
    \includegraphics[width=1.0\textwidth]{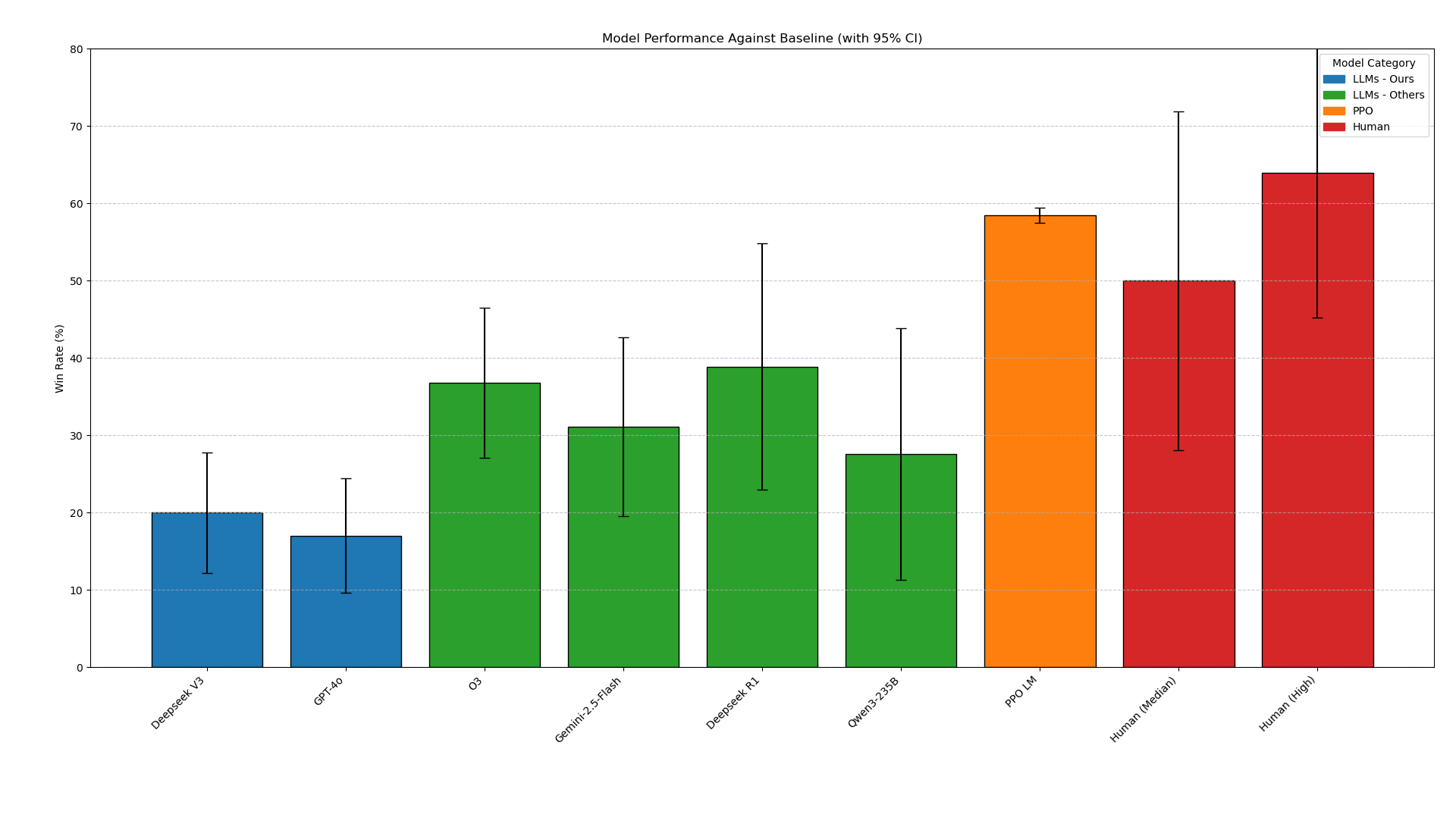} 
    \caption{Visualization of Model Performance Metrics (Win Rates against Baseline). Please replace this placeholder with figure1.png.}
    \label{fig:performance_visualization}
\end{figure}

The PPO-based LM achieved the highest win rate of $58.5\% \pm 1.0\%$ over 10,000 games, significantly outperforming all LLM agents. Among the LLMs, DeepSeek R1 ($38.9\% \pm 15.9\%$) and O3 ($36.8\% \pm 9.7\%$) showed relatively stronger performance, while others like GPT-4o ($17.0\% \pm 7.4\%$) struggled more against the baseline. The performance of the "Human Expert (High)" ($64.0\% \pm 18.8\%$), representing STEM students who have studied the game, sets a strong benchmark that the PPO agent approaches. The "Human Expert (Median)" ($50.0\% \pm 21.9\%$) represents proficient players. Figure \ref{fig:performance_visualization} (placeholder) is intended to visually compare these win rates.
Further analysis of the PPO agent's training would ideally include its learning curve, plotting win rate against the baseline or average reward per episode over training steps/epochs. This would illustrate the agent's learning progress and convergence. Additionally, a breakdown of game statistics, such as average game length when playing with/against different agents, or the average number of correct/incorrect guesses per game, could provide deeper insights into their respective playing styles and efficiencies. For LLMs, a qualitative analysis of common error types (e.g., rule violations versus suboptimal strategic choices) would also be valuable, potentially linking these to the complexity of the prompt or the inherent limitations of the LLM architecture for this task. Computational resources for training the PPO agent involved standard GPU hardware over several days, while LLM agent evaluations were API call-dependent.

\section{Discussion}
The superior performance of the PPO-based agent ($58.5\%$ win rate) compared to the LLM agents (best at $38.9\%$) and the baseline Transformer underscores the efficacy of deep reinforcement learning, when coupled with a history-aware state representation like a Transformer encoder, for mastering deductive games such as Da Vinci Code. The PPO agent's capacity to learn a policy directly from extensive self-play allows it to implicitly discover and refine complex deductive strategies, manage uncertainty, and optimize for long-term rewards in a way that is challenging for prompted LLMs or simpler supervised models. The agent's processing of full game history is crucial, enabling it to track revealed information, infer constraints from past actions (both its own and the opponent's), and dynamically update its belief state about hidden tiles.

LLMs, despite their advanced reasoning capabilities and the provision of detailed prompts with game rules and strategic heuristics, faced notable challenges. The observation that some LLMs produced outputs inconsistent with game logic (as noted in Table \ref{tab:model_performance}) points to difficulties in strictly adhering to formal rule systems over extended interactions or in complex combinatorial settings. This behavior aligns with broader findings where LLMs may struggle with multi-step symbolic reasoning or precise state tracking without more specialized architectures (e.g., external memory) or fine-tuning on domain-specific data. The "median candidate" heuristic in the Gemini prompt, while a logical starting point, likely simplifies the true strategic depth of Da Vinci Code. Expert human play often involves more nuanced probabilistic assessments, opponent modeling (e.g., inferring an opponent's risk aversion from their guessing patterns), and even elements of bluffing or strategic misdirection concerning one's own revealed tiles, all of which are aspects an RL agent might implicitly learn over millions of games but are hard to encapsulate in a fixed prompt.

The challenges related to policy optimization and behavior regulation highlighted in research on other complex card games like Guandan \citep{pan2024mastering} and DouDizhu \citep{zha2021douzero} are relevant here. While our PPO agent demonstrates strong performance, scaling such an approach to games with even vaster combinatorial action spaces or more intricate multi-agent dynamics (e.g., cooperative play within teams) would likely demand further innovations in state abstraction, action parameterization for large discrete or continuous spaces, and more sophisticated exploration strategies. However, Da Vinci Code's core emphasis on logical deduction based on a growing set of public facts might make its state more amenable to structured representations and its optimal policies more learnable compared to games dominated by stochastic card distributions and complex hand-play combinations. The PPO framework's relative sample efficiency and stability make it a good candidate for such tasks, but fine-tuning the reward structure and state representation remains critical.

Comparing the developmental costs, the PPO agent required significant computational resources for training (many GPU hours over days) and careful design of the state/action spaces and reward function. LLM agents, conversely, were faster to set up for evaluation (primarily involving prompt engineering and API calls) but incurred per-inference costs and exhibited less reliable performance. This trade-off between upfront training investment for specialized agents versus the flexibility and rapid prototyping potential of generalist LLMs is a key consideration in AI development for games.

\section{Project Scope and Future Work}
The project successfully implemented a Da Vinci Code game environment and several AI agents: a baseline Transformer, LLM-based agents (Gemini, DeepSeek, etc.), and a PPO-trained agent. A GUI for Transformer vs. human play was also partially developed.
Key tasks realized include:
\begin{enumerate}
    \item \textbf{Game Environment:} A robust \texttt{DaVinciCodeGameEnvironment} was fully implemented, serving as the foundation for all agent development and evaluation.
    \item \textbf{PPO Agent Training:} The \texttt{test\_ppo.py} script establishes a complete training pipeline for the PPO agent, including state encoding, network updates, and experience replay.
    \item \textbf{LLM Prompting and Evaluation:} The \texttt{gemini.py} script showcases the design of prompts for LLM agents and a framework for their interaction with the game environment.
\end{enumerate}
Future work could explore several promising avenues to build upon these findings:
\begin{itemize}
    \item \textbf{Enhanced LLM Reasoning and Grounding:} Investigate methods to improve LLM logical consistency and strategic planning. This could involve advanced prompting techniques like Chain-of-Thought or Tree-of-Thoughts \citep{yao2023tree}, mechanisms for self-correction based on rule violations, or fine-tuning LLMs on a corpus of Da Vinci Code game data (expert games or PPO agent trajectories) to better ground their reasoning in the game's specific logic.
    \item \textbf{Hybrid AI Approaches:} Explore synergistic combinations of RL and LLMs. For instance, an LLM could be used for high-level strategic guidance (e.g., identifying promising lines of deduction or assessing overall game phase) that informs the policy of an RL agent. Conversely, an RL agent could generate a diverse set of candidate actions, which an LLM then evaluates or refines based on deeper logical scrutiny.
    \item \textbf{Sophisticated Opponent Modeling:} Explicitly modeling opponent tendencies, risk profiles, or common deductive errors could further enhance the PPO agent's performance, allowing it to exploit predictable patterns or adapt to different opponent styles, especially crucial in multi-round or tournament settings.
    \item \textbf{Scalability and Generalization Studies:} Test the robustness and adaptability of the PPO agent framework on more complex deduction games (e.g., Clue/Cluedo, Mastermind variants with more dimensions) or games with significantly larger state/action spaces to understand its scaling limitations and potential for generalization.
    \item \textbf{Ablation Studies and Architectural Refinements:} Conduct thorough ablation studies on the PPO agent's components (e.g., impact of history length in state encoding, different Transformer configurations, alternative reward shaping functions) to pinpoint key drivers of performance and guide further architectural refinements.
    \item \textbf{Explainable AI (XAI) for Deductive Agents:} Develop methods to interpret the PPO agent's decision-making process, potentially visualizing its attention weights or extracting human-understandable deductive rules from its learned policy. This would not only build trust but also offer insights into novel strategies the AI might discover.
\end{itemize}

\section{Conclusion}
This paper presented a comprehensive comparative study of Transformer-based, LLM-based, and PPO-based AI agents for the deductive game of Da Vinci Code. Our findings demonstrate that a PPO agent, equipped with a Transformer encoder to process rich game history, significantly outperforms both a history-limited Transformer baseline and several state-of-the-art LLMs guided by sophisticated prompts. The PPO agent achieved a win rate of 58.5\% against the baseline, approaching the proficiency of skilled human players who have studied the game. While LLMs exhibited deductive reasoning capabilities, their performance was hampered by challenges in maintaining strict logical consistency and developing deep strategic acumen through prompting alone. This research underscores the substantial potential of tailored deep reinforcement learning approaches for mastering complex deductive games characterized by imperfect information. The detailed implementation and comparative analysis provide a valuable framework and empirical insights for future investigations in AI for games requiring multi-step logical inference and strategic planning.

\bibliography{references} 

\appendix
\section{Technical Appendices}
Further details on network architectures, hyperparameters, or ablation studies can be placed here.

\subsection{Da Vinci Code Gameplay Algorithm}
\label{app:gameplay_algorithm}
The core gameplay loop for Da Vinci Code is outlined in Algorithm \ref{alg:davinci_code_flow}.

\begin{algorithm}[H] 
\caption{Da Vinci Code Game Flow}
\label{alg:davinci_code_flow}
\begin{algorithmic}[1] 
\STATE \textbf{Game Setup}:
\STATE Shuffle all number tiles (Black and White 0--11) and two joker/hyphen tiles.
\STATE Each player draws tiles:
    \begin{itemize}[label=\textbullet, leftmargin=*]
        \item 2--3 players: 4 tiles each.
        \item 4 players: 3 tiles each.
    \end{itemize}
\STATE Players arrange their drawn tiles vertically in front of them, hidden from opponents (opponents see the back of the tiles), according to the following rules:
    \begin{itemize}[label=\textbullet, leftmargin=*]
        \item Tiles are sorted by number in ascending order.
        \item For tiles with the same number, Black tiles are placed to the left of White tiles.
    \end{itemize}
\STATE Remaining tiles are placed face down to form the draw pile.

\STATE \textbf{Game Process}:
\WHILE{game is not over}
    \FOR{each player in clockwise order}
        \STATE Draw one tile from the pile (if available); view its content privately.
        \REPEAT
            \STATE Select an opponent and one of their unrevealed tiles; guess its color and number.
            \IF{guess is correct}
                \STATE The guessed opponent reveals the tile; it remains in its position.
                \STATE Current player may choose to guess another unrevealed tile or end their current guessing phase.
            \ELSE[guess is incorrect]
                \STATE Current player reveals the tile they just drew (if one was drawn this turn) and inserts it into their tile row according to ordering rules. If no tile was drawn, they reveal one of their existing hidden tiles.
                \STATE End current player's turn.
            \ENDIF
        \UNTIL{guess is incorrect OR player chooses to end guessing phase (after a correct guess)}
        \IF{player chose to end guessing phase after a correct guess AND a tile was drawn this turn}
            \STATE Player places the drawn tile into their row (hidden).
            \STATE End current player's turn.
        \ENDIF
    \ENDFOR
\ENDWHILE

\STATE \textbf{Game End Condition}:
\STATE When only one player has unrevealed tiles remaining, that player wins.
\end{algorithmic}
\end{algorithm}

\subsection{GUI for Human-Agent Interaction}
\begin{figure}[H] 
    \centering
    \includegraphics[width=1.0\textwidth]{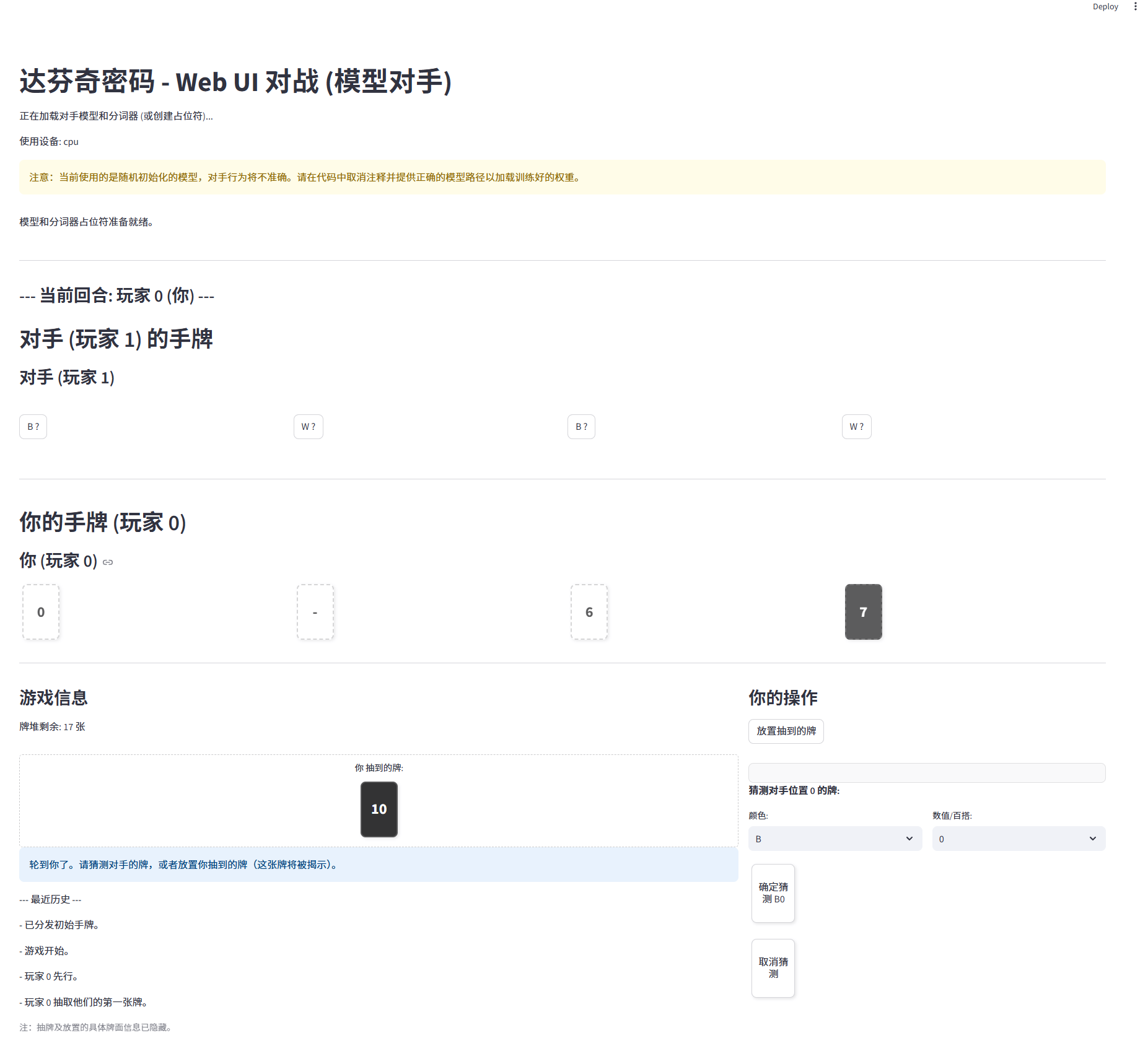} 
    \caption{GUI for Human-AI Da Vinci Code Gameplay. Please replace this placeholder with figure\_2.png.}
    \label{fig:gui_interface}
\end{figure}

Figure \ref{fig:gui_interface} shows a placeholder for the Graphical User Interface (GUI) developed for human players to compete against the AI agents. This interface facilitates game setup, displays the game state, and allows for human input of actions.

\end{document}